\documentclass[preprint,3p,review,12pt]{elsarticle}

\usepackage{amssymb}

\usepackage{lineno}
\modulolinenumbers[5]

\usepackage{amsmath,amsfonts}
\usepackage{algorithmic}
\usepackage{algorithm}
\usepackage{array}
\usepackage[caption=false,font=normalsize,labelfont=sf,textfont=sf]{subfig}
\usepackage{textcomp}
\usepackage{stfloats}
\usepackage{url}
\usepackage{verbatim}
\usepackage{graphicx}

\usepackage{framed,multirow}
\usepackage{latexsym}
\usepackage{color}
\usepackage{siunitx}
\usepackage{dsfont}
\usepackage{booktabs}
\usepackage{makecell}
\usepackage{tabularx}
\usepackage{bbding}
\usepackage{tipx}
\usepackage{hyperref}
\hypersetup{hypertex=true,
            colorlinks=true,
            linkcolor=red,
            anchorcolor=blue,
            citecolor=blue}

\newcommand{\etal}{\textit{et al.}}

\newcolumntype{A}{>{\color{blue}}l}
\newcolumntype{B}{>{\color{blue}}c}
\newcolumntype{C}{>{\color{blue}}r}

\journal{Pattern Recognition}

\begin{document}


\begin{frontmatter}



\title{Class-Aware Mask-Guided Feature Refinement\\ for Scene Text Recognition}

\author[label1]{Mingkun Yang}
\ead{yangmingkun@hust.edu.cn}

\author[label2]{Biao Yang}
\ead{hust\_byang@hust.edu.cn}

\author[label5]{Minghui Liao}
\ead{mhliao@foxmail.com}

\author[label1,label3]{Yingying Zhu\corref{cor1}}
\cortext[cor1]{Corresponding author.}
\ead{yyzhu@hust.edu.cn}

\author[label4]{Xiang Bai}
\ead{xbai@hust.edu.cn}

\affiliation[label1]{organization={School of Electronic Information and Communications, Huazhong University of Science and Technology},
            city={Wuhan},
            country={China}}
\affiliation[label2]{organization={School of Artificial Intelligence and Automation, Huazhong University of Science and Technology},
            city={Wuhan},
            country={China}}
\affiliation[label3]{organization={Hubei Key Laboratory of Smart Internet Technology, Huazhong University of Science and Technology},
            city={Wuhan},
            country={China}}
\affiliation[label4]{organization={School of Software Engineering, Huazhong University of Science and Technology},
            city={Wuhan},
            country={China}}
\affiliation[label5]{organization={Huawei Inc.},
            city={Shenzhen},
            country={China}}

\begin{abstract}

Scene text recognition is a rapidly developing field that faces numerous challenges due to the complexity and diversity of scene text, including complex backgrounds, diverse fonts, flexible arrangements, and accidental occlusions. In this paper, we propose a novel approach called Class-Aware Mask-guided feature refinement (CAM) to address these challenges. Our approach introduces canonical class-aware glyph masks generated from a standard font to effectively suppress background and text style noise, thereby enhancing feature discrimination. Additionally, we design a feature alignment and fusion module to incorporate the canonical mask guidance for further feature refinement for text recognition. By enhancing the alignment between the canonical mask feature and the text feature, the module ensures more effective fusion, ultimately leading to improved recognition performance. We first evaluate CAM on six standard text recognition benchmarks to demonstrate its effectiveness. Furthermore, CAM exhibits superiority over the state-of-the-art method by an average performance gain of \textbf{4.1\%} across six more challenging datasets, despite utilizing a smaller model size. Our study highlights the importance of incorporating canonical mask guidance and aligned feature refinement techniques for robust scene text recognition. The code is available at \url{https://github.com/MelosY/CAM}.

\end{abstract}

\begin{keyword}
Text Recognition, Text Segmentation, Multi-task Learning, Feature Fusion
\end{keyword}

\end{frontmatter}



\section{Introduction}

    \begin{figure}[h]
        \centering
        \includegraphics[width=1.0\linewidth]{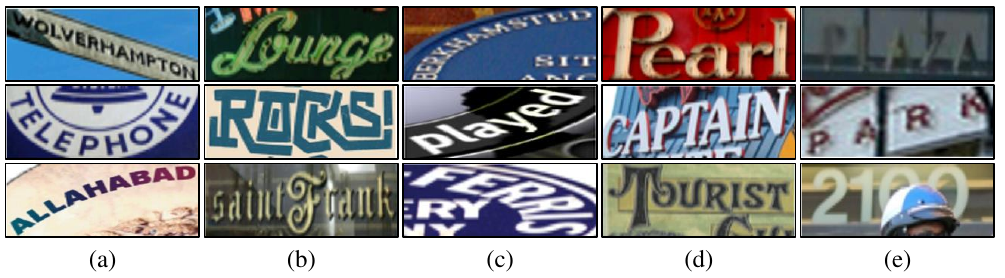}
        \vspace{-7ex}
        \caption{Text images with (a) various layouts, (b) diverse fonts, (c) perspective distortion, (d) clutter backgrounds, and (e) occlusion.}
        \label{fig:challenging_imgs}
    \end{figure}

    Scene text recognition (STR) is a pivotal task that involves the automatic recognition of text within natural images, such as street signs~\cite{zhu2017cascaded}, billboards~\cite{conf/icdar/ZhangYBSKLJZJSL19}, and product labels~\cite{tip/LiaoSB18}. The applications of STR span across numerous domains, including industrial automation, image-based geo-location, document analysis~\cite{vidal2023end}, and human-computer interaction. The remarkable progress in computer vision and natural language processing has fueled the rapid development of scene text recognition in recent years. Leveraging sophisticated deep learning architectures, large-scale annotated datasets, and algorithmic innovations, state-of-the-art STR methods continue to push the boundaries of accuracy and robustness, paving the way for even more advanced applications and advancements in this dynamic field.

    Due to the inherent challenges posed by complex backgrounds, diverse fonts, flexible arrangements, and accidental occlusions in scene text, as illustrated in Fig.~\ref{fig:challenging_imgs}, STR remains a demanding task, especially in these difficult scenarios. Previous methods have attempted to address these challenges by incorporating related tasks into the text recognition framework to harness additional information for improving recognition performance. Notably, some methods utilized Generative Adversarial Networks (GANs) or segmentation networks to aid in text recognition. For example, Luo~\etal~\cite{journals/ijcv/LuoLLJS21} employed GANs to alleviate the complexity of the background in text images. Liu~\etal~\cite{conf/eccv/LiuWJW18} and Wang~\etal~\cite{conf/mm/WangL20} proposed multi-task frameworks that integrated text recognition and class-agnostic glyph mask generation via GANs. Another approach, introduced by Zhao~\etal~\cite{BINet}, incorporated a text segmentation task and utilized the obtained segmentation information to enhance the features fed to the text recognition decoder. However, most of these methods implicitly impact text recognition through only sharing the encoder, which may limit the effective utilization of valuable information from the GAN. Although Zhao~\etal~\cite{BINet} explicitly integrates segmentation information into the text recognition decoder, it necessitates pixel-level annotations and an independent segmentation network for foreground/background prediction, which incurs heavy annotation and computation costs. 
    These limitations motivate the need for a novel approach that can leverage the benefits of GANs or segmentation networks explicitly within the STR framework while minimizing the reliance on extensive manual annotations and additional network components. More detailed comparisons are discussed in Sec.~\ref{subsec:relatedwork_mask}.

    In this paper, we present a novel approach, \textbf{C}lass-\textbf{A}ware \textbf{M}ask-guided feature refinement (CAM), specifically tailored to tackle the aforementioned challenges in STR. 
    CAM leverages prototypes to refine recognition features, where the prototype takes the form of a canonical class-aware glyph mask generated using a standard font. This mask furnishes clear and consistent information for each character class to enhance feature discrimination.
    Diverging from prior methods~\cite{conf/eccv/LiuWJW18,journals/ijcv/LuoLLJS21,BINet,conf/mm/WangL20}, which employed class-agnostic masks as targets or even pixel-level annotations~\cite{BINet}, CAM incorporates a class-aware canonical glyph segmentation module to enhance feature discrimination. To avoid solely relying on a shared encoder to implicitly impact text recognition and effectively fuse unaligned features, we introduce a mask-guided feature alignment and fusion module. This module explicitly enhances the features for text recognition, promoting efficient alignment and fusion. By combining prototype guidance and feature refinement through alignment and fusion, our method bolsters the discrimination of the learned features, thereby advancing the overall performance of text recognition.

    The contributions of this paper are summarized as follows:
    \begin{enumerate}
    	\item We propose a feature refinement approach that utilizes a prototype to guide the feature learning process. The prototype, in the form of a canonical glyph mask with a standard font, effectively highlights the text while eliminating style and background noise.
    
    	\item We introduce a discriminative canonical glyph segmentation module that improves feature discrimination compared to previous class-agnostic GAN or segmentation-based methods. It enhances the ability to distinguish different characters by incorporating class-specific information.
    
    	\item We design a mask-guided feature alignment and fusion module to explicitly enhance the features for text recognition. This module effectively alleviates the pixel-level misalignment between the features extracted from input with arbitrary shapes and the corresponding canonical glyph mask, which further improves recognition performance by complementary fusion.
    
    	\item Our proposed method achieves state-of-the-art results on standard scene text recognition benchmarks. Moreover, it outperforms the previous best method by an average of \textbf{4.1\%} across six challenging datasets, demonstrating its superiority in handling complex and diverse scene text recognition scenarios.
    \end{enumerate}

\section{Related works}
    Inspired by the advancements in automatic speech recognition (ASR) and machine translation, STR is usually viewed as a sequence generation task in recent years. Based on the decoder types for sequence generation, STR methods can be divided into the following categories: CTC-based~\cite{conf/icpr/XuZBCNP20,crnn,conf/aaai/HeH0LT16,conf/ijcnn/GaoZL21}, attention-based~\cite{aster,YangGLHBBYB19,focus_attn,conf/aaai/WangZJLCWWC20}, and transformer-based~\cite{conf/mm/YangLLWZLTB22,wordart,DBLP:journals/pr/LinLJL21,DBLP:journals/pr/LuYQCGXB21} methods. Specifically, CTC-based methods, while simple and efficient, are limited in capturing contextual information. Attention-based methods leverage recurrent neural networks (RNNs) to understand text semantics but suffer from time-by-time decoding, which can impede recognition efficiency. Transformer-based decoders strike a balance between efficiency and performance by allowing parallel character generation during training and iterative generation during inference. Besides the above methods that rely on a predefined vocabulary, there have been efforts to address more open scenarios through various approaches such as semi-supervised learning~\cite{gao2021semi}, self-supervised learning~\cite{conf/mm/YangLLWZLTB22}, open-set learning~\cite{liu2023towards}, and others. More related works are referred to \cite{journals/fcsc/ZhuYB16,journals/ijcv/LongHY21}. Considering the extensive body of literature on STR, we only discuss works most closely related to our approach, such as irregular text recognition and mask-related text recognition.
    
\subsection{Irregular Text Recognition}
    Because of diverse layouts and various perspectives of text instances, the captured scene text images are always oriented, perspectively distorted, or curved, which greatly increases the difficulty of recognition.
    
    A common solution is to rectify text images into regular forms and then apply a conventional recognition algorithm~\cite{aster,journals/tmm/Wu0HCL23,journals/ijon/WangL21,DBLP:journals/pr/LuoJS19}. The pioneering framework ASTER~\cite{aster} introduced an end-to-end framework for irregular STR, which employed Thin Plate Spline (TPS) to explicitly rectify distorted and curved texts before recognition.
    To further improve the rectification effect, 
    ScRN~\cite{YangGLHBBYB19} introduced text geometric attributes to fit the TPS transformation, resulting in a more accurate rectification effect. Recently, TPS++~\cite{zheng2023tps++} built a more flexible content-aware rectifier via a joint process of foreground control point regression and content-based attention score estimation. An alternative solution for recognizing irregular text involves a two-dimensional (2D) perspective. \cite{show_attend_read} adopted a 2D attention-based encoder-decoder network. AON~\cite{aon} encoded the input image into four feature sequences representing different directions to handle oriented text instances.
    Liao~\etal~\cite{LiaoZWXLLYB19} formulated text recognition of arbitrary shapes as a semantic segmentation task; however, this approach requires character-level annotations for label generation, limiting its practical application.
    
    Generally, rectification-based methods often yield improvable rectification results due to the implicit supervision from text recognition, while 2D-based methods excel in handling highly distorted text but with higher computational costs. We aim to improve basic rectification-based methods without imposing a significant extra computation burden.

\subsection{Mask-related Text Recognition}
\label{subsec:relatedwork_mask}
    Besides dealing with irregular arrangements, methods for STR also face challenges related to text instances with diverse styles and complex backgrounds. Some existing works have attempted to address these issues by removing background noise or font styles to enhance scene text recognition.
    
    One such approach is employed by Luo~\etal~\cite{journals/ijcv/LuoLLJS21}, who utilized Generative Adversarial Networks (GANs) to separate text content from cluttered backgrounds. By reducing the background noise, the recognition of text on the generated images becomes easier. However, because of the serial pipeline, the recognition performance heavily relies on the quality of the generated features from GAN.
    Another technique, known as BINet~\cite{BINet}, addresses the issue of noisy backgrounds by employing an individual segmentation network pre-trained on two real-world datasets~\cite{textseg,total_text} with pixel-level annotations.
    To explore font-independent features for scene text, Liu~\etal~\cite{conf/eccv/LiuWJW18} proposed a GAN-based approach that transforms the entire scene text image into corresponding horizontally written canonical masks. To ensure stable training, feature-level guidance from clean images to input images is utilized, resulting in increased computational overhead. Furthermore, the guidance is implicitly imposed solely on the encoder through backward propagation, which may not be the optimal solution for text recognition.
    In contrast to generating entire text images, Wang~\etal~\cite{conf/mm/WangL20} generated canonical characters through attentional generation of glyphs in numerous font styles. However, this method overlooks word-level layouts and lacks explicit complementation between recognition and generation.
    
    Motivated by the aforementioned analyses, we aim to present a unified framework capable of handling cluttered backgrounds, diverse font styles, and irregular layouts simultaneously, while minimizing the computational requirements. Instead of employing the class-agnostic binary segmentation, we introduce class-aware canonical mask-guided learning to facilitate discriminative feature learning. Additionally, we go beyond the conventional practice of implicitly incorporating the information of canonical glyph masks through backward propagation. Instead, we explicitly fuse the original recognition features with the canonical features and propose a novel module that effectively mitigates the pixel-level misalignment between them.

\section{Methodology}
\label{sec:methodology}

    \begin{figure*}[h]
        \centering
        \includegraphics[width=0.95\linewidth]{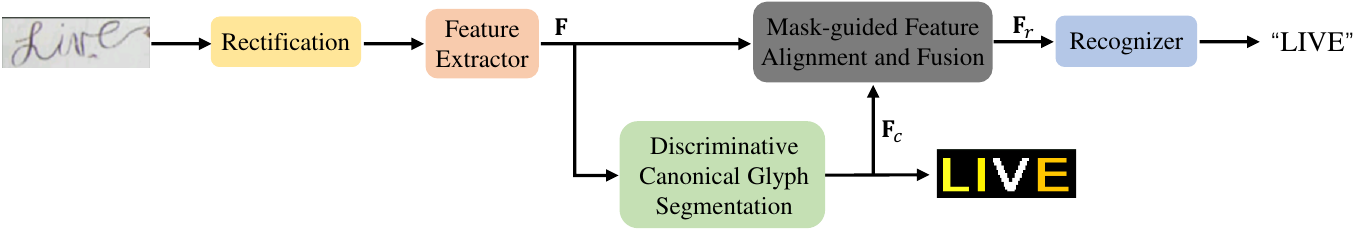}
        \vspace{-2ex}
        \caption{Illustration of our proposed CAM. $\mathbf{F}$ is the visual feature extracted from the backbone. $\mathbf{F}_c$ is the embedding of canonical glyph masks. $\mathbf{F}_r$ is the fused feature of canonical masks and backbone features.}
        \vspace{-2ex}
        \label{fig:pipeline}
    \end{figure*}

    Text recognition endeavors to convert images containing text in various shapes, fonts, and intricate backgrounds into uniform character sequences devoid of visual properties. To bridge the gap between diverse images and uniform labels, we introduce canonical class-aware glyphs to the text recognition method. The overall architecture of our model is illustrated in Fig.~\ref{fig:pipeline}, comprising an image transformation module, a feature extractor, a discriminative canonical glyph segmentation module, a mask-guided feature alignment and fusion module, and a transformer-based decoder.

\subsection{Image Rectification and Feature Extraction}
\label{subsec:imgrec_featenc}

    First, the image rectification module employs a TPS-based STN to transform the input image $\mathbf{I}$ into a normalized image $\mathbf{I}_r\in\mathbb{R}^{3 \times H \times W}$, with dimensions of 32 and 128 for height ($H$) and width ($W$), respectively. Although the rectification results may not always be sufficient in challenging scenarios due to the TPS transformation parameters being learned implicitly via recognition, the coarse rectification still provides a suitable input for subsequent canonical glyph learning. For a more detailed mathematical derivation of the TPS transformation, we recommend referring to the works RARE~\cite{rare} and ASTER~\cite{aster}.
 
    Recently, using Vision Transformer (ViT) \cite{transformer} for feature extraction has become increasingly popular in STR methods~\cite{wordart,conf/eccv/BautistaA22parseq,conf/eccv/DaWY22levocr,conf/eccv/WangDY22multigra}. Because of the capability to capture long-range dependencies, ViT demonstrates superiority in various computer vision tasks while incurring significant computation complexity. To compete favorably with ViT while retaining the simplicity and efficiency of standard ConvNet, we employ ConvNeXt V2 \cite{woo2023convnext} as our feature extractor. By modifying the original strides to $(4,4)$, $(2,1)$, $(2,1)$, $(1,1)$, we ensure a suitable trade-off between computation complexity and the capability for subsequent spatial segmentation. As a result, the size of output feature $\mathbf{F}$ is 2$\times$32 for an input image of size 32$\times$128. Moreover, to compete with existing STR methods in various model sizes, we also construct multiple variants of ConvNeXt V2 by adjusting the channels and the number of blocks in each stage.

\subsection{Discriminative Canonical Glyph Segmentation}
\label{subsec:segmentation}

    \begin{figure}[t]
        \centering
        \includegraphics[width=0.5\linewidth]{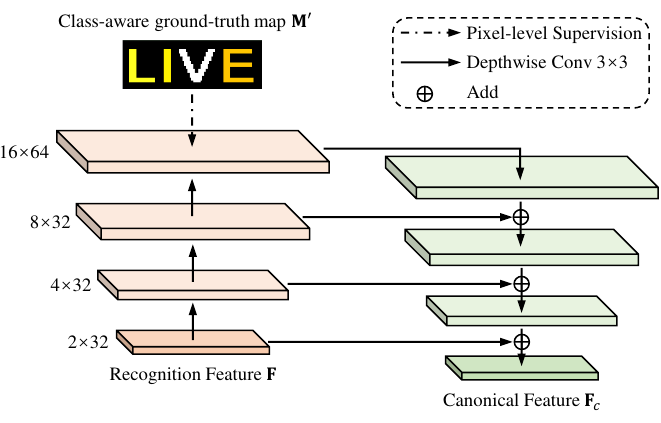}
        \vspace{-3ex}
        \caption{Illustration of the discriminative canonical glyph segmentation module, comprising the class-aware segmentation component (left) and a canonical feature generation structure with a ``$\cap$-shaped" design.}
        \label{fig:segmentation}
    \end{figure}

    We introduce class-aware canonical glyphs to enhance text recognition performance, particularly in the presence of diverse layouts, varied font styles, and cluttered backgrounds. These canonical glyphs are automatically generated through a synthetic engine, leveraging existing ground-truth labels and requiring no additional annotations. As depicted in Fig.~\ref{fig:segmentation}, we retain category information to incorporate more discriminative visual information into the feature extractor. Ultimately, we obtain the ground-truth map $\mathbf{M}^{'}\in\mathbb{R}^{H \times W}$, where the value of each pixel denotes its character type.

    After the rectification and feature extraction stages, we obtain a coarsely normalized feature map $\mathbf{F}\in\mathbb{R}^{C \times \frac{H}{16} \times \frac{W}{4}}$. Subsequently, a lightweight segmentation module with depthwise convolution is constructed to perform pixel-level prediction, as outlined in Tab.~\ref{tab:arch_seg}. Finally, we reshape the resulting prediction to $\mathbf{M}\in\mathbb{R}^{(N_c+1) \times H \times W}$, where $N_c$ represents the number of character categories, and the remaining category corresponds to the background.

    In addition to predicting canonical glyphs, this module generates an embedding $\mathbf{F}_c\in\mathbb{R}^{C \times \frac{H}{16} \times \frac{W}{4}}$ of canonical glyphs to enhance the recognition features. The architecture, illustrated in Fig.~\ref{fig:segmentation}, is inspired by the popular segmentation method UNet~\cite{ronneberger2015u}. We introduce a reversed U-shaped structure that effectively captures high-level features while preserving fine details. The module comprises an upsampling decoder and a downsampling encoder, connected through a ``$\cap$-shaped" skip connection. The downsampling decoder follows a similar architecture to the upsampling encoder described in Tab.~\ref{tab:arch_seg}, with the only difference being the replacement of Up-Sample layers with downsampling convolutional layers. The skip connection is established by adding the output of each layer in the decoder network with the corresponding layer in the encoder network.

    \begin{table}[!t]
        \newcommand{\tabincell}[2]{\begin{tabular}{@{}#1@{}}#2\end{tabular}}
        \centering
        \caption{The architecture of segmentation module. The configuration has the following format: {$\{kernel_h \times kernel_w, stride_h \times stride_w, pad_h \times pad_w, channels\}$} for convolutional layers, \{dimensions\} for the number of features in fully-connected layers. ``Output Size" is the feature map size of convolutional layers or output dimension of fully-connected layers.}
        \begin{center}
            \resizebox{0.55\textwidth}{!}{
            \begin{tabular}{clc}
            \toprule
            Layer Name  & \multicolumn{1}{c}{Configuration}      & Output Size \cr\midrule
        
            \multirow{2}{*}{Stage1}
            & DepthConv: $3\times3, 1\times1, 1\times1, C$ & \multirow{2}{*}{$\frac{H}{16}\times \frac{W}{4}$} \cr\cmidrule{2-2}
            & BatchNorm &   \cr\midrule
            
            \multirow{5}{*}{Stage2}
            & Up-Sample: $4\times1, 2\times1, 1\times0, C$ & \multirow{5}{*}{$\frac{H}{8}\times \frac{W}{4}$} \cr\cmidrule{2-2}
            & DepthConv: $3\times3, 1\times1, 1\times1, C$ &  \cr\cmidrule{2-2}
            & BatchNorm, ReLU & \cr\cmidrule{2-2}
            & DepthConv: $3\times3, 1\times1, 1\times1, C/2$ &  \cr\cmidrule{2-2}
            & BatchNorm & \cr\midrule
            
            \multirow{5}{*}{Stage3}
            & Up-Sample: $4\times1, 2\times1, 1\times0, C/2$ & \multirow{5}{*}{$\frac{H}{4}\times \frac{W}{4}$} \cr\cmidrule{2-2}
            & DepthConv: $3\times3, 1\times1, 1\times1, C/2$ &  \cr\cmidrule{2-2}
            & BatchNorm, ReLU & \cr\cmidrule{2-2}
            & DepthConv: $3\times3, 1\times1, 1\times1, C/4$ &  \cr\cmidrule{2-2}
            & BatchNorm & \cr\midrule
            
            \multirow{5}{*}{Stage4}
            & Up-Sample: $4\times4, 2\times2, 1\times1, C/4$ & \multirow{5}{*}{$\frac{H}{2}\times \frac{W}{2}$} \cr\cmidrule{2-2}
            & DepthConv: $3\times3, 1\times1, 1\times1, C/4$ &  \cr\cmidrule{2-2}
            & BatchNorm, ReLU & \cr\cmidrule{2-2}
            & DepthConv: $3\times3, 1\times1, 1\times1, C/8$ &  \cr\cmidrule{2-2}
            & BatchNorm & \cr\midrule
            
            FC & $2\times2\times (N_c+1)$ & $2\times2\times (N_c+1)$ \cr\bottomrule
            \end{tabular}}
        \end{center}
        \label{tab:arch_seg}
    \end{table}

\subsection{Mask-guided Feature Alignment and Fusion}
\label{subsec:refinement}

    The generated canonical class-aware masks from the segmentation module are generally easier to recognize due to the absence of distortion factors. However, to prevent cumulative errors, we do not directly utilize the predicted mask for recognition. Instead, we employ it to compensate for the original recognition features.

    We note that the canonical mask does not inherently ensure pixel-level alignment with the recognition feature. As a result, conventional pixel-level fusion operations, such as \textit{Add}, \textit{Dot Product}, \textit{Concatenate}, and \textit{Conditional Normalization} \cite{BINet}, may not be suitable for our task. To overcome this limitation, we introduce a mask-guided aligned feature fusion module. Inspired by \cite{conf/cvpr/XiaPSLH22}, where dynamic region selection for conventional self-attention is proposed, our approach effectively aligns the two features by learning flexible deformed sampling points on the recognition features in a mask-dependent manner. Subsequently, the aligned features are fused to leverage their complementary information.

    \begin{figure*}[t]
        \centering
        \includegraphics[width=1.\linewidth]{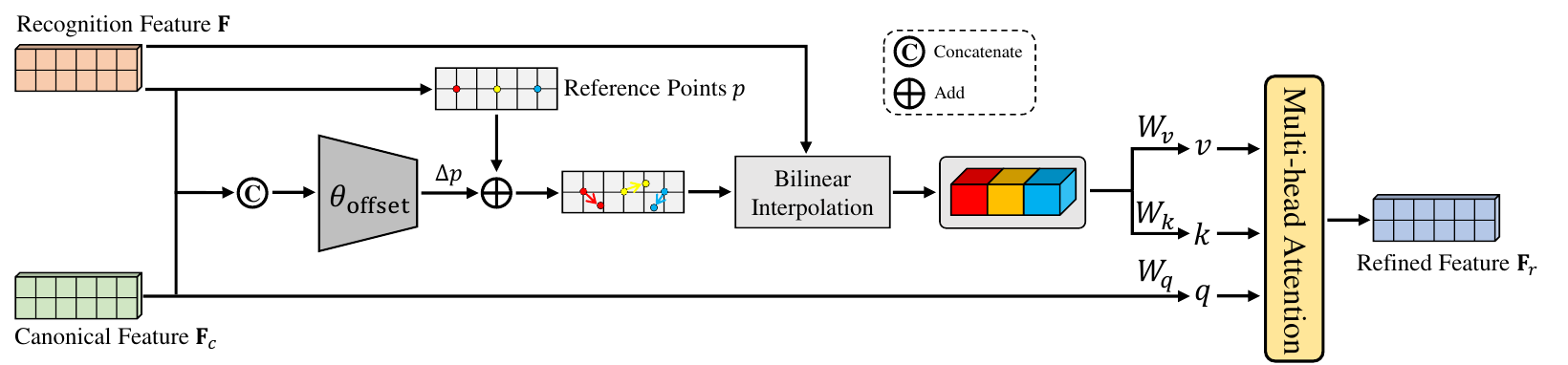}
        \caption{Illustration of the mask-guided feature alignment and fusion module. Firstly, a set of reference points is placed uniformly on the recognition feature maps, whose offsets are learned from the concatenated features of recognition features and canonical features. Subsequently, the deformed keys and values are projected from the sampled features according to the deformed points. Finally, regular multi-head attention is employed for fusion.}
        \label{fig:deform_fusion}
    \end{figure*}
    
    The architecture is illustrated in Fig.~\ref{fig:deform_fusion}. The recognition feature $\mathbf{F}$ and canonical feature $\mathbf{F}_{c}$ are aligned through deformed sampling points, which are generated by dynamically predicting a set of offsets $\Delta p$ based on pre-defined reference points $p$. The reference points are generated as a uniform grid of points $p \in \mathbb{R}^{H_g\times{}W_g\times{}2}$, where the grid size is downsampled from the recognition feature map size by a factor of $r$. Specifically, $H_g$ is set as $(H/16)/r$ and $W_g$ as $(W/4)/r$. The 2D coordinates of the points on the grid are linearly spaced between $(0,0)$ and $(H_g-1, W_g-1)$. Furthermore, we normalize the 2D coordinates of the reference points to the range of $[-1,+1]$, based on the grid shape $H_g\times{}W_g$.

    The offsets to the reference points are learned to dynamically align $\mathbf{F}$ and $\mathbf{F}_c$. To achieve this, we concatenate $\mathbf{F}$ and $\mathbf{F}_c$ and pass them through a two-layer convolutional block $\theta_\text{offset}$ to generate the offsets $\Delta{}p=\theta_\text{offset}(\text{concat}(\mathbf{F},\mathbf{F}_c))$. Additionally, to diversify the deformed points, we split the feature channels into $G$ groups and employ a shared sub-network to generate groups of offsets.
    
    Next, the aligned feature $\tilde{\mathbf{F}}$ is obtained by sampling from the recognition feature $\mathbf{F}$ at the location of the deformed points using bilinear interpolation $\phi$. Subsequently, the aligned features $\tilde{\mathbf{F}}$ serve as keys and values, while the canonical feature $\mathbf{F}_c$ acts as queries, followed by projection matrices:
    \begin{align}
        q=&{\mathbf{F}_c}W_q,\ k=\tilde{\mathbf{F}}W_k,\ v=\tilde{\mathbf{F}}W_v, \label{eq:proj_dmha} \\
        \text{with} \ & \ \tilde{\mathbf{F}}=\phi(\mathbf{F};p+\Delta{}p). \label{eq:sampling}
    \end{align}
    \noindent where $W_q$, $W_k$, $W_v$ are learnable parameters.

    Following the original transformer layer, we utilize multi-head attention on the queries $q$, keys $k$, and values $v$. The output of each attention head is computed as follows,
    \begin{equation}
        \mathbf{F}_r^{(m)}=\text{softmax}\left(q^{(m)}{k}^{(m)\top}/\sqrt{d}\right){v}^{(m)}.
    \end{equation}
    where $d$ is the dimension of each attention head. Then, the features from each attention head are concatenated and projected using a linear layer to obtain the final output $\mathbf{F}_r$.

\subsection{Recognition Decoder}
    Finally, a two-layer transformer decoder is employed to generate the final prediction. Each transformer layer consists of three core modules: Masked Multi-Head Self Attention, Multi-Head Cross Attention, and Feed Forward Network. The Masked Multi-Head Self Attention ensures that the prediction at the current step only utilizes information from its previous steps. During training, the decoder predicts all steps simultaneously by employing a lower triangle mask, enabling highly parallel training and significantly reducing training time compared to traditional step-by-step decoding. The Multi-Head Cross Attention follows a similar formulation as the vanilla self-attention layer but with queries originating from the previous decoder layer and keys and values sourced from the encoder's output. This mechanism allows every position in the decoder to attend to all positions in the input sequence, resembling the encoder-decoder attention commonly seen in sequence-to-sequence models.

\subsection{Objective}
\label{subsec:overall_obj}

    The overall objective function is formulated as a weighted sum of losses for text recognition and pixel-wise segmentation:

    \begin{equation}
        \mathcal{L} = \mathcal{L}_{rec} + \lambda\mathcal{L}_{seg},
    \label{eqn:overall_obj}
    \end{equation}

    \noindent where $\lambda$ is empirically set to 1.0. The $\mathcal{L}_{rec}$ and $\mathcal{L}_{seg}$ are computed as follows.

    Given the refined feature $\mathbf{F}_r$ and its ground-truth character sequence $\mathbf{Y}=\{y_1,y_2,...,y_T\}$, the transformer decoder produces a sequence of probability distribution $\mathbf{P}=\{p_1,p_2,...,p_T\}$ normalized via \textit{Softmax}. The recognition model is optimized by minimizing the negative log-likelihood of the conditional probability as follows:

    \begin{equation}
        \mathcal{L}_{rec} = - \frac{1}{T}\sum_i^{T}\log p_i(y_i \mid \mathbf{I})
    \label{eqn:loss_ccr}
    \end{equation}

    We employ a weighted cross-entropy loss to train the class-aware segmentation branch. Let $\mathbf{M}_{c,i,j}$ represent an element of the prediction map, where $i \in \{1, ..., H\}$, $j \in \{1, ..., W\}$, and $c \in \{0,1, ..., N_c\}$. Furthermore, let $\mathbf{M}^{'}_{i,j} \in \{0,1, ..., N_c\}$ denote the corresponding class label. The prediction loss can be calculated as follows,
    
    \begin{equation}
        \begin{aligned}
            \mathcal{L}_{seg} = & -\frac{1}{H \times W}\sum_{i=1}^{H}\sum_{j=1}^{W}w_{i,j}(\\
            & \sum_{c=0}^{N_c} 
            \mathds{1}(\mathbf{M}^{'}_{i,j}==c)\log(\frac{e^{\mathbf{M}_{c,i,j}}}{\sum_{k=0}^{N_c} e^{\mathbf{M}_{c,i,j}}})),
        \end{aligned}
    \end{equation}
    where $\mathds{1}(\cdot)$ is an indicator function. $w_{i,j}$ is the corresponding weight of each pixel. Assume that $N=H \times W$ and $N_{neg}$ is the number of background pixels. The weight can be calculated as follows,
    
    \begin{equation}
         w_{i,j} = 
        \begin{cases}
            N_{neg} / (N - N_{neg})& \text{if } \mathbf{M}^{'}_{i,j}>0, \\
            1& \text{otherwise}
        \end{cases}
    \end{equation}

    Through joint optimization, the feature extractor is encouraged to complement the visual features with fine details, which further facilitates text recognition.

\section{Experiments}
\subsection{Datasets}
    Our proposed CAM are evaluated on both English and Chinese datasets.
    For English, all models are trained using two widely used synthetic datasets, namely SynthText~\cite{synthtext} and Synth90k~\cite{synth90k}, which consist of approximately 8 million and 9 million images, respectively. As for the evaluation, besides the standard benchmarks, we also conduct assessments on more challenging datasets that are closer to practical real-world scenarios. These datasets encompass a diverse range of scenarios, exhibiting significant variations in fonts, sizes, orientations, and backgrounds. Moreover, these benchmarks incorporate images with low-quality attributes, such as noise, blur, distortion, or occlusion. Consequently, the inclusion of these challenging datasets allows us to thoroughly assess the robustness and effectiveness of our STR models in recognizing diverse and complex text within natural environments. Further details can be found in Tab.~\ref{tab:benchmarks}.

    For Chinese, we have identified a comprehensive dataset for Chinese text recognition in~\cite{yu2021benchmarking}, which provides an in-depth analysis with some representative text recognition methods. The dataset is divided into four distinct subsets based on different scenarios: \textit{Scene}, \textit{Web}, \textit{Document}, and \textit{Handwriting}. For each scenario, we train the proposed CAM on the corresponding training set and employ the validation set to select the optimal model. Subsequently, the selected model is evaluated on the respective test set.
    
    \begin{table}[t]
        \centering
        \caption{Details of Scene Text Recognition Benchmarks.}
        \label{tab:benchmarks}
        \resizebox{0.6\textwidth}{!}{
        \begin{tabular}{lcc}
        \toprule
        Benchmarks                                 & \# Test images    & Description \\ \midrule
        IIIT~\cite{conf/cvpr/MishraAJ12}      &   3000            &  Regular Scene Text           \\ 
        SVT~\cite{wang2011end}                     &   647             &  Regular Scene Text          \\ 
        IC13~\cite{karatzas2013icdar}              &   1015            &  Regular Scene Text          \\ \midrule
        IC15~\cite{karatzas2015icdar}              &   1811            &  Incidental Scene Text          \\
        COCO~\cite{cocotext}                       &   9896            &  Incidental Scene Text          \\ \midrule
        SVTP~\cite{quy2013recognizing}             &   645             &  Perspective Scene Text         \\ \midrule
        CUTE~\cite{risnumawan2014robust}           &   288             &  Curved Scene Text         \\ 
        CTW~\cite{ctw1500}                         &   1572            &  Curved Scene Text         \\ 
        TT~\cite{total_text}                       &   2201            &  Curved Scene Text        \\ \midrule
        WOST~\cite{ost}                            &   2416            &  Weakly Occluded Scene Text         \\ 
        HOST~\cite{ost}                            &   2416            &  Heavily Occluded Scene Text         \\ \midrule
        WordArt~\cite{wordart}                     &   1511            &  Scene Text of wordart    \\
        \bottomrule
        \end{tabular}}
    \end{table}

\subsection{Implementation Details}

    \subsubsection{Image Rectification} For the image rectification module, we follow the settings of ASTER~\cite{aster}. To be consistent with the feature extractor and segmentation module, the convolution layers are replaced with depthwise convolution layers. The rectification sampling is performed on the input image of size $64\times 256$, resulting in a transformed image of size $32\times 128$ for the subsequent recognition. To predict the control points required for image rectification, we employ a lightweight network that operates on the downsampled input image of size $32\times 64$. This network consists of six convolutional layers, each utilizing a $3\times 3$ kernel. Each of the first five convolutional layers is followed by a $2\times 2$ max-pooling layer. The output filters for the convolutional layers are set to 32, 64, 128, 256, 256, and 256, respectively. Two fully-connected layers follow the convolutional layers, with 512 and 2K units as their respective output sizes. Here, K represents the number of control points. In our implementation, we set 10 control points on both the upper and lower edges, resulting in a total of 20 control points.

    \subsubsection{Feature Extractor} As the feature encoder, we utilize a vanilla ConvNext V2 model. For the decoder, we employ a two-layer transformer block with embedding dimension 384 and eight heads. To ensure comparability with various state-of-the-art text recognizers that employ different model sizes, we incorporate three ConvNext V2 variants: ConvNext V2-Nano, ConvNext V2-Tiny, and ConvNext V2-Base. The architectural details of these variants are consistent with those described in \cite{woo2023convnext}. Furthermore, the decoder and segmentation module are consistently scaled up along with the feature extractor. To reduce experimental overhead, we adopt ConvNext V2-Nano as the default backbone for the ablation study. Consequently, the final recognition models are named CAM-Nano, CAM-Tiny, and CAM-Base, corresponding to the different ConvNext V2 variants.

    \subsubsection{Ground-truth Labels} For text recognition and class-aware segmentation, our model considers a total of 68 character categories, including digits, lowercase letters, and 32 punctuation marks. Moreover, the text recognition task includes an additional symbol denoting the end-of-sequence (EOS), while the class-aware segmentation task involves an extra background category. To ensure clarity in pixel-level segmentation, our framework treats letters as case-insensitive, eliminating any pixel-level semantic ambiguity.

    \subsubsection{Optimization} We utilize AdamW optimizer and a \textit{cosine} learning rate scheduler. For English models, the training hyperparameters are set as follows: batch size of 128, base learning rate of 8e-4, weight decay of 0.05, $\beta_1$ = 0.9, $\beta_2$ = 0.999, and a warm-up period of one epoch, with a total of 6 epochs. However, as the model size increases, the learning rate and batch size are slightly adjusted to accommodate limited memory and ensure a stable training process. Specifically, the initial learning rates for CAM-Tiny and CAM-Base are set to 4e-4 and 2e-4, respectively. The batch size for CAM-Base is reduced to 64. All experiments are conducted using 8 NVIDIA P100 (16GB RAM) GPUs.

    In alignment with the experiments on English datasets, most hyperparameters for Chinese models remain identical. Following the approach in~\cite{conf/ijcai/DuCJYZLDJ22svtr}, we set the epoch to 100 for the \textit{Scene} subset. Considering the varying sizes of each sub-dataset, we adjust the epoch values for \textit{Web}, \textit{Document}, and \textit{Handwriting} to 500, 127, and 700, respectively. The maximum prediction length is set to 40. Furthermore, the vocabulary sizes are individually set to 5883, 4405, 4868, and 6108, corresponding to each of the four subsets. Due to the limited GPU memory, the batch size on each GPU for CAM-Nano, CAM-Tiny, and CAM-Base is reduced to 64, 64, and 32, respectively. Additionally, we set the learning rate to 4e-4 for the \textit{Web} and \textit{Document} subsets, ensuring stability during the training process.

    \subsubsection{Data Augmentation} We incorporate commonly used data augmentation techniques to enhance the robustness and generalization of our STR model. Following \cite{conf/eccv/BautistaA22parseq}, we apply a set of image augmentations, including perspective distortion, affine distortion, blur, noise, and rotation in the training stage.
    
\subsection{Comparison with State-of-the-Art Methods}
    \subsubsection{Recognition on Common Datasets}
    
\begin{table*}[!t]
\centering
\caption{Scene text recognition results on standard benchmarks, categorized as ``V" for language-free models and ``VL" for language-aware models. The best results are presented in bold font, while underlined values indicate the second-best results.}
\label{tab:sota_common}
\resizebox{0.975\textwidth}{!}{
\begin{tabular}{@{}l|c|c|cccccc|c|c|c@{}}
\toprule
\multicolumn{1}{c|}{Methods} & Types & Venue      & IIIT          & SVT           & IC13          & IC15          & SVTP          & CUTE          & Avg.          & Params (M) & Times (ms)\\ \midrule
TRBA~\cite{baek2019wrong}    & V     & CVPR2021   & 92.1          & 88.9          & 93.9          & 78.3          & 79.5          & 78.2          & 87.1          & 50         & 18.1 \\
PREN2D~\cite{Pren2D}         & V     & CVPR2021   & 95.6          & 94.0          & 96.4          & 83.0          & 87.6          & 91.7          & 91.6          & -          & - \\
SATRN~\cite{satrn}           & V     & CVPRW2021  & 92.8          & 91.3          & -             & -             & 86.5          & 87.8          & -             & 55         & 265.5 \\
Text is Text~\cite{textistext} & V     & ICCV2021   & 92.3          & 89.9          & 93.3          & 76.9          & 84.4          & 86.3          & 87.5          & -        & - \\
Mask TextSpotter~\cite{LiaoLHYWB21} & V     & TPAMI2021  & 95.3          & 91.8          & 95.3          & 78.2          & 83.6          & 88.5          & 89.5          & -   & -       \\
ConCLR~\cite{conf/aaai/ZhangZYSL022conclr} & V     & AAAI2022   & 95.7          & 92.1          & 95.9          & 84.4          & 85.7          & 89.2          & 91.5   & -   & -       \\
SVTR-L~\cite{conf/ijcai/DuCJYZLDJ22svtr} & V     & IJCAI2022  & 96.3          & 91.7          & 97.2          & 86.6          & 88.4          & \textbf{95.1} & 92.9     & 41  & 21.9       \\
CornerTransformer~\cite{wordart} & V     & ECCV2022   & 95.9          & 94.6          & 96.4          & 86.3          & \textbf{91.5} & 92.0          & 93.0          & 86     & 212.1  \\
$\text{SIGA}_R$~\cite{guan2023self} & V     & CVPR2023   & 95.9          & 92.7          & 97.0          & 85.1          & 87.1          & 91.7          & 92.2          & 40  & 104.7   \\
$\text{SIGA}_T$~\cite{guan2023self} & V     & CVPR2023   & 96.6          & \underline{95.1}          & \textbf{97.8} & 86.6          & 90.5          & \underline{93.1}      & 93.5          & 113    & 133.5    \\ \midrule
PIMNet~\cite{pimnet}                  & VL    & ACM MM2021 & 95.2          & 91.2          & 95.2          & 83.5          & 84.3          & 84.4          & 90.6          & - & -        \\
ABINet~\cite{abinet}         & VL    & CVPR2021   & 96.2          & 93.5          & 97.4          & 86.0          & 89.3          & 89.2          & 92.8          & 37         & 23.4 \\
VisionLAN~\cite{ost}         & VL    & ICCV2021   & 95.8          & 91.7          & 95.7          & 83.7          & 86.0          & 88.5          & 91.3          & 33         & 17.6 \\
JVSR~\cite{JVSR}             & VL    & ICCV2021   & 95.2          & 92.2          & -             & -             & 85.7          & 89.7          & -             & -          & -\\
SGBANet~\cite{Sgbanet}       & VL    & ECCV2022   & 95.4          & 89.1          & 95.1          & 78.4          & 83.1          & 88.2          & 89.3          & -          & -\\
LevOCR~\cite{conf/eccv/DaWY22levocr} & VL    & ECCV2022   & 96.6          & 92.9          & 96.7          & 86.4          & 88.1          & 91.7          & 92.9          & 93 & 120.7       \\
$\text{MGP-STR}_{Fuse}$~\cite{conf/eccv/WangDY22multigra} & VL    & ECCV2022   & 96.4          & 94.7          & 97.3          & \underline{87.2}          & \underline{91.0}          & 90.3          & 93.4          & 148  & 14.9     \\
$\text{PARSeq}_A$~\cite{conf/eccv/BautistaA22parseq}                       & VL    & ECCV2022   & \underline{97.0}          & 93.6          & 96.2          & 82.9          & 88.9          & 92.2          & 92.3          & 24   & 13.8     \\ \midrule
CAM-Nano                    & V     & -          & \textbf{97.3}          & 94.6          & 96.7          & 87.0          & 88.8          & 90.6          & \textbf{93.5}          & 23       & 15.3  \\
CAM-Tiny                    & V     & -          & \textbf{97.3}          & \textbf{96.0}          & \underline{97.7} & \textbf{87.5}          & 90.7          & 92.4          & \textbf{94.1} & 58   &21.2      \\
CAM-Base                    & V     & -          & \textbf{97.4} & \textbf{96.1} & 97.2          & \textbf{87.8} & 90.6          & 92.4          & \textbf{94.1} & 135  &26.8      \\ \bottomrule
\end{tabular}}
\end{table*}

    Following most existing methods, we compare the proposed method against SOTA approaches on six commonly used benchmarks. The results, presented in Tab.~\ref{tab:sota_common}, showcase the superiority of CAM across a range of model sizes, from the low-capacity 23M CAM-Nano model to the high-capacity 135M CAM-Base model.
    
    \textbf{Comparison with Language-free Models.} Remarkably, CAM-Nano (23M) achieves the best average performance while utilizing the fewest parameters. Scaling up the model to CAM-Tiny and CAM-Base further enhances the final performance, achieving an impressive accuracy of 94.1\%. When compared to the most recent method $\text{SIGA}_T$, CAM outperforms it in four out of six cases. Specifically, CAM-Tiny demonstrates superior performance to $\text{SIGA}_T$ on IIIT, SVT, IC15, and SVTP by 0.8\%, 1.0\%, 1.2\%, and 0.1\%, respectively. Furthermore, our results on IC13 and CUTE are highly comparable to the SOTA, with CAM-Tiny only being 0.1\% lower than $\text{SIGA}_T$ on IC13 and making only two additional error predictions on CUTE.
    
    \textbf{Comparison with Language-aware Models.} Recently, language-aware methods have gained popularity in the STR field by explicitly incorporating linguistic priors to improve initial text predictions. These methods often achieve satisfactory performance on benchmarks with rich contextual information. We further compare CAM with similar-sized language-aware approaches. Surprisingly, CAM achieves the best accuracy across all six benchmarks without employing explicit language modeling. Specifically, CAM-Nano outperforms PARSeq on IIIT, SVT, IC13, and IC15 by 0.3\%, 1.0\%, 0.5\%, and 4.1\%, respectively. Additionally, CAM-Tiny outperforms $\text{MGP-STR}_{Fuse}$ on IIIT, SVT, IC13, IC15, and CUTE by 0.9\%, 1.3\%, 0.4\%, 0.3\%, and 2.1\%, respectively.

    Furthermore, we introduce the inference time as another crucial criterion to comprehensively evaluate these methods. To ensure a fair comparison, we gather the publicly released codes of some recent methods and reevaluate their inference steps on a single Nvidia V100 GPU. The inference time, averaged over 1000 images, is utilized for comparison. The results are summarized in Table~\ref{tab:sota_common}. We find that $\text{MGP-STR}_{Fuse}$ employs a non-autoregressive decoder, and $\text{PARSeq}_A$ adopts an 8×4 patch with fewer tokens in ViT, which enables them to achieve slightly faster inference times compared to CAM. However, CAM-Nano achieves the best performance while maintaining a highly comparable inference time. Consequently, we believe that CAM strikes a better balance between performance, parameter efficiency, and runtime.
    
    In conclusion, our robust visual feature learning approach demonstrates its effectiveness by surpassing state-of-the-art and language-aware methods on most standard benchmarks. These results underline the importance of further exploration and development of robust visual feature learning techniques in the field of STR.
    
    \subsubsection{Recognition on Challenging Datasets}

    With the advancements in computer vision and natural language processing, modern STR methods have exhibited outstanding performance in a wide range of practical applications. However, due to the limited diversity and scale of commonly used benchmarks, they may not provide comprehensive insights into the performance of current methods. To address this limitation, we conduct comparative experiments on six challenging datasets that offer greater complexity. To ensure fairness and transparency in our evaluation, we select several SOTA works that have publicly available codes and models for comparison. The results, as depicted in Tab.~\ref{tab:sota_challenging}, demonstrate the efficacy of our proposed method in tackling the challenges posed by these datasets.

    \textbf{Comparison with Language-free Models.} Consequently, CAM shows a significant superiority across these demanding benchmarks. When compared to the most recent method $\text{SIGA}_T$, CAM-Tiny achieves an average improvement of \textbf{4.1\%} and exhibits performance enhancements on these datasets by 7.5\%, 0.2\%, 1.2\%, 1.4\%, 1.0\%, and 0.2\%, respectively. Notably, the substantial performance gain of 7.5\% on COCO highlights the robustness of our proposed class-aware mask in more natural scenarios compared to the self-supervised implicit glyph attention.

\begin{table*}[]
\centering
\caption{Comparison results on challenging datasets.}
\label{tab:sota_challenging}
\resizebox{0.95\textwidth}{!}{
\begin{tabular}{@{}l|c|c|cccccc|c|c@{}}
\toprule
\multicolumn{1}{c|}{Methods} & Types & Venue    & COCO          & CTW           & TT            & HOST          & WOST          & WordArt       & Avg.          & Params (M) \\ \midrule
CornerTransformer~\cite{wordart}            & V     & ECCV2022 & 63.2          & 77.5          & 81.4          & 64.7          & 77.4          & 70.8          & 68.8          & 86         \\
$\text{SIGA}_R$~\cite{guan2023self}              & V     & CVPR2023 & 63.2          & 76.8          & 80.5          & 56.2          & 75.6          & 69.9          & 67.3          & 40         \\
$\text{SIGA}_T$~\cite{guan2023self}              & V     & CVPR2023 & 62.0          & \underline{79.3}          & \underline{82.2}          & \underline{77.2}          & \underline{85.1}          & \underline{74.9}          & \underline{71.2}          & 113        \\ \midrule
LevOCR~\cite{conf/eccv/DaWY22levocr}                       & VL    & ECCV2022 & 64.5          & 77.8          & 79.9          & 51.3          & 74.1          & 71.0          & 67.3          & 93         \\
$\text{MGP-STR}_{Fuse}$~\cite{conf/eccv/WangDY22multigra}      & VL    & ECCV2022 & \underline{65.3}          & 76.7          & 80.4          & 67.8          & 81.0          & 72.4          & 70.6          & 148        \\
$\text{PARSeq}_A$~\cite{conf/eccv/BautistaA22parseq}                       & VL    & ECCV2022 & 64.4          & 78.6          & 80.1          & 73.1          & 81.6          & 72.5          & 71.0          & 24         \\ \midrule
CAM-Nano                    & V     & -        & \textbf{68.7} & 78.4          & 82.1          & 76.2          & \textbf{85.1} & 72.9          & \textbf{74.1} & 23         \\
CAM-Tiny                    & V     & -        & \textbf{69.5} & \textbf{79.5} & \textbf{83.4} & \textbf{78.6} & \textbf{86.1} & \textbf{75.1} & \textbf{75.3} & 58         \\
CAM-Base                    & V     & -        & \textbf{69.7} & \textbf{80.1} & \textbf{83.6} & \textbf{78.2} & \textbf{86.9} & \textbf{75.6} & \textbf{75.6} & 135        \\ \bottomrule
\end{tabular}}
\end{table*}

    \textbf{Comparison with Language-aware Models.} Among the SOTA methods, PARSeq can achieve nearly the best performance while utilizing the fewest parameters. CAM-Nano, with a comparable parameter size, outperforms PARSeq in five out of six cases. Notably, CAM-Nano achieves impressive performance gains of 4.3\%, 2.0\%, 3.1\%, 3.5\%, and 0.5\% on COCO, TT, HOST, WOST, and WordArt, respectively, resulting in an average improvement of \textbf{3.1\%}.

    By evaluating CAM on these more demanding benchmarks, we are able to showcase its robustness in various real-world scenarios. The significant improvements on these challenging datasets further validate the effectiveness of our approach.

    \subsubsection{Recognition on Chinese Datasets}

    To further validate the effectiveness and generalization of our proposed method, we evaluate CAM on the Chinese datasets. The results, as shown in Table~\ref{tab:Chinese_res}, provide further confirmation of the superiority of our approach. Specifically, our CAM-Nano surpasses the state-of-the-art performance by 2.33\%, 6.3\%, and 4.51\% on the \textit{Scene}, \textit{Web}, and \textit{Handwriting} subsets, respectively. Moreover, CAM-Nano, CAM-Tiny, and CAM-Base achieve an average performance gain of 3.91\%, 5.32\%, and 6.56\% compared to TransOCR. The results convincingly demonstrate the remarkable cross-language and cross-scenario generalization ability of CAM.

    In addition, we observe an interesting phenomenon that CAM-Base generally outperforms CAM-Tiny on 9 out of 12 benchmarks, while CAM-Tiny with fewer parameters exhibits better performance on IC13, SVTP, and HOST by 0.5\%, 0.1\%, and 0.4\%, respectively. Following the standard practice in existing English STR methods, CAM is trained using synthetic images. With the same experimental settings, CAM-Tiny has achieved state-of-the-art average performance, suggesting that further improvements on these benchmarks may be limited. Moreover, the diversity inherent in the synthetic images may be somewhat modest, and thus CAM-Tiny (58M) could be sufficient to effectively model the distribution of the synthetic training data. However, CAM-Base, with significantly more parameters (135M), potentially faces the risk of overfitting to these synthetic images, resulting in slightly lower performance on 3 out of 12 benchmarks.
    In Tab~\ref{tab:Chinese_res}, both the training and evaluation of CAM are conducted using real-world Chinese datasets. The results demonstrate that when the training data exhibits extensive diversity, the larger model can achieve notable performance gains. Specifically, CAM-Base outperforms CAM-Tiny by 2.0\%, 0.5\%, 0.2\%, and 1.9\%, respectively. Thus, we believe that the diversity of the training dataset could significantly influence the scalability of CAM.

    \begin{table*}[t]
    \centering
    \caption{Comparison with previous methods on Chinese benchmarks. The first eight results are derived from \cite{yu2021benchmarking}.}
    \resizebox{0.75\textwidth}{!}{
    \begin{tabular}{l|c|cccc|c}
    \toprule
    Method & Venue & Scene & Web & Document & Handwriting & Avg. \\
    \midrule
    CRNN~\cite{crnn} & TPAMI2017 & 54.9 & 56.2 & 97.4 & 48.0 & 68.0 \\
    ASTER~\cite{aster} & TPAMI2019 & 59.4 & 57.8 & 91.6 & 45.9 & 67.8 \\
    MORAN~\cite{DBLP:journals/pr/LuoJS19} & PR2019 & 54.7 & 49.6 & 91.7 & 30.2 & 62.7 \\ 
    SAR~\cite{show_attend_read} & AAAI2019 & 53.8 & 50.5 & 96.2 & 31.0 & 64.0 \\
    SEED~\cite{SE_ASTER} & CVPR2020 & 45.4 & 31.4 & 96.1 & 21.1 & 57.1 \\
    MASTER~\cite{master} & PR2021 & 62.1 & 53.4 & 82.7 & 18.5 & 61.4 \\
    ABINet~\cite{abinet} & CVPR2021 & 60.9 & 51.1 & 91.7 & 13.8 & 62.9 \\
    TransOCR~\cite{Chen2021SceneTT} & CVPR2021 & 67.8 & \underline{62.7} & \underline{97.9} & \underline{51.7} & \underline{74.8} \\
    SVTR-L~\cite{conf/ijcai/DuCJYZLDJ22svtr} & IJCAI2022 & \underline{72.1} & - & - & - & -  \\
    \midrule
    CAM-Nano & - & \textbf{74.3} & \textbf{69.0} & 97.5 & \textbf{56.2} & \textbf{78.7} \\
    CAM-Tiny & - & \textbf{76.0} & \textbf{69.3} & \textbf{98.1} & \textbf{59.2} & \textbf{80.1} \\
    CAM-Base & - & \textbf{78.0} & \textbf{69.8} & \textbf{98.3} & \textbf{61.1} & \textbf{81.3} \\
    \bottomrule
    \end{tabular}}
    \label{tab:Chinese_res}
    \end{table*}
    
\subsection{Ablation Study}

    \begin{table}[]
    \centering
    \begin{center}
    \caption{Ablation study on model units. The terms ``Class-agnostic Mask" and ``Class-aware Mask" denote the usage of the canonical binary mask and the canonical class-aware mask for segmentation supervision, respectively.}
    \label{tab:ablation}
    \scalebox{0.75}{
    \begin{tabular}{ccc|cccccc|c}
    \toprule
    \multicolumn{1}{c}{\begin{tabular}[c]{@{}c@{}}Class-agnostic\\ Mask\end{tabular}} & \begin{tabular}[c]{@{}c@{}}Class-aware\\ Mask\end{tabular} & \begin{tabular}[c]{@{}c@{}}Aligned\\ Fusion\end{tabular} & Regular & Incidental & Perspective & Curved & Occluded & WordArt & Avg. \\ \hline
                                                                                      &                                                            &                                                          & \underline{96.5}    & \underline{71.0}       & \underline{88.1}        & 79.4   & 76.6     & 71.3    & 77.8 \\ 
                                                                \CheckmarkBold        &                                                            &                                                          & 96.3    & 70.8       & 87.6        & 79.8   & 77.3     & 72.0    & 78.0 \\ 
                                                                                      &                    \CheckmarkBold                          &                                                          & 96.4    & 70.8       & \textbf{88.8}        & \underline{80.2}   & \underline{79.4}     & \underline{72.6}    & \underline{78.5} \\ 
                                                                                      &                    \CheckmarkBold                          &            \CheckmarkBold                                & \textbf{96.8}    & \textbf{71.5}       & \textbf{88.8}        & \textbf{81.3}   & \textbf{80.7}     & \textbf{72.9}    & \textbf{79.3} \\ \bottomrule
    \end{tabular}}
    \end{center}
    \end{table}

    In this section, we analyze two pivotal components, namely the class-aware mask segmentation and mask-guided aligned feature fusion, which are co-designed to synergistically enhance the text recognizer. We begin by examining the individual contributions of each module unit towards the overall recognition performance. Subsequently, we delve deeper into the detailed designs within each unit.
    
    \subsubsection{Effectiveness of model units}
    To clearly demonstrate the effectiveness of each module unit in specific scenarios, we present the average performance based on the dataset type, as shown in Tab.~\ref{tab:benchmarks}.
    
    \textbf{Class-agnostic Mask \textit{vs.} Class-aware Mask.} In this section, we add the extra segmentation branch to recognition for multi-task learning. In previous works, a canonical binary mask or binary foreground/background mask was used to simplify text recognition. Thus, it is a natural choice to incorporate such a class-agnostic mask into our framework. As presented in Tab.~\ref{tab:ablation}, the class-agnostic mask achieves performance improvements of 0.4\%, 0.7\%, and 0.7\% on the \textit{Curved}, \textit{Occluded}, and \textit{Wordart} datasets, respectively, with an average improvement of 0.2\%. However, its performance is slightly lower, yet comparable, on the other datasets. By integrating our proposed class-aware mask, the recognizer outperforms the class-agnostic variant in five out of six scenarios, exhibiting accuracy improvements of 0.7\%, 0.8\%, 2.8\%, and 1.3\% on the \textit{Perspective}, \textit{Curved}, \textit{Occluded}, and \textit{Wordart} datasets, respectively. We believe that the canonical mask (both class-agnostic and class-aware masks) effectively normalizes the input images in terms of irregular layouts, various fonts, and occasional occlusions, resulting in performance gains on the corresponding datasets. Moreover, the class-aware mask carries specific category information, contributing more discriminative features to the visual feature extractor through the segmentation network.
    
    \textbf{Superiority of Aligned Fusion.} Furthermore, when combined with the introduced fusion module, our STR model exhibits performance improvements of 0.4\%, 0.7\%, 1.1\%, 1.3\%, and 0.3\% on the \textit{Regular}, \textit{Incidental}, \textit{Curved}, \textit{Occluded}, and \textit{Wordart} datasets, respectively. The notable enhancements on the \textit{Curved} and \textit{Occluded} datasets indicate that the mask-guided alignment and fusion module not only rectifies irregular text at the feature level but also provides complementary features for occluded text.

    Ultimately, our model surpasses the baseline model with significant improvements of 0.3\%, 0.5\%, 0.7\%, 1.9\%, 4.1\%, and 1.6\% across all dataset types, yielding an average improvement of 1.5\%. These experimental results convincingly demonstrate the advantages of our proposed modules. Particularly noteworthy are the substantial enhancements achieved on the \textit{Curved}, \textit{Occluded}, and \textit{Wordart} datasets, affirming the effectiveness of our method in aligning irregular text to canonical layouts, compensating for incomplete text, and discriminating varied fonts. Furthermore, we observe that our proposed modules facilitate faster convergence during the training process. We speculate that the utilization of class-aware masks, exhibiting canonical layouts, consistent styles, clean backgrounds, and pixel-level discriminative supervision, can effectively bridge the gap between diverse text images and uniform label sequences. This intermediate representation provides explicit and simplified guidance, thereby accelerating the convergence of the training process.

    \subsubsection{Discussion on fusion strategies}

    As another crucial component, the fusion of recognition features and canonical segmentation features involves different strategies. In addition to the widely-used methods such as \textit{Add}, \textit{Dot Product}, and \textit{Concatenate}, we also compare our fusion methods with those employed in the field of STR, including \textit{Conditional Normalization}~\cite{BINet} and \textit{Cross Attention}~\cite{wordart}. The results of these fusion strategies are presented in Tab.~\ref{tab:ablation_fusion}. We can observe that the previous fusion methods yield improvements of 0.1\% - 0.2\% compared to the variant without fusion, while our designed alignment-based fusion module outperforms these competitors by 0.6\% - 0.7\%. The underlying reason can be attributed to the fact that in \cite{BINet}, the segmentation map of the foreground is accurately aligned with the recognition features at a pixel level. Consequently, pixel-wise operations such as addition, multiplication, concatenation, and conditional normalization yield satisfactory performance. However, in the absence of such detailed manual segmentation annotations, the synthesized canonical mask cannot guarantee pixel-level alignment, which may even impair the recognition feature if using pixel-level fusion.

\begin{table}
    \centering
    \begin{center}
        \caption{Comparison between different fusion strategies.}
        \label{tab:ablation_fusion}
        \scalebox{0.725}{
            \begin{tabular}{c|cccccccccccc|c}
            \toprule
                \begin{tabular}[c]{@{}c@{}}Fusion\\ Strategy\end{tabular}           & IIIT          & SVT           & IC13          & IC15          & SVTP          & CUTE          & COCO          & CTW           & TT            & HOST          & WOST          & WordArt       & Avg.          \\ \midrule
                Add                                                                 & {\underline{97.2}}    & 93.7          & {\underline{96.6}}    & 86.0          & {\underline{88.4}}    & 89.2          & \textbf{68.8} & 77.5          & 80.7          & 74.5          & 83.1          & {\underline{72.6}}    & 78.6          \\
                Dot Product                                                         & 96.9          & 93.8          & \textbf{96.7} & 86.5          & 87.8          & 89.2          & 68.6          & 77.0          & 81.3          & {\underline{75.4}}    & 83.9          & 72.2          & {\underline{78.7}}    \\
                Concatenate                                                         & 96.9          & 93.7          & {\underline{96.6}}    & \textbf{87.3} & {\underline{88.4}}    & 88.9          & 68.4          & 77.4          & 81.2          & 75.0          & 83.7          & 72.3          & {\underline{78.7}}    \\
                \begin{tabular}[c]{@{}c@{}}Conditional\\ Normalization\end{tabular} & {\underline{97.2}}    & {\underline{94.0}}    & 96.1          & 86.7          & {\underline{88.4}}    & {\underline{89.9}}    & 68.1          & 77.4          & {\underline{81.6}}    & 74.1          & {\underline{84.6}}    & 72.3          & 78.6          \\
                \begin{tabular}[c]{@{}c@{}}Cross\\ Attention\end{tabular}           & {\underline{97.2}}    & {\underline{94.0}}    & 96.5          & 86.9          & 88.1          & \textbf{90.6} & 68.4          & {\underline{77.6}}    & 81.3          & 74.3          & 84.4          & 72.3          & {\underline{78.7}}    \\
                 \makecell{Aligned \\ Fusion}            & \textbf{97.3} & \textbf{94.6} & \textbf{96.7} & {\underline{87.0}}    & \textbf{88.8} & \textbf{90.6} & {\underline{68.7}}    & \textbf{78.4} & \textbf{82.1} & \textbf{76.2} & \textbf{85.1} & \textbf{72.9} & \textbf{79.3} \\ \bottomrule
            \end{tabular}
        }
    \end{center}
\end{table}

    Transformer-based cross-attention becomes a natural choice to employ dense attention for feature selection and further aligned fusion. We can find that the aligned fusion outperforms the cross-attention fusion on almost all datasets. Specifically, the aligned fusion achieves improvements of 0.6\%, 0.7\%, 0.8\%, 0.8\%, 1.9\%, 0.7\%, 0.6\% on SVT, SVTP, CTW, TT, HOST, WOST, and WordArt, respectively. We think the global receptive field in the cross-attention can influence features by irrelevant parts that lie beyond the region of interest. In our proposed fusion module, we address this limitation by enabling dynamic selection of key and value pairs in a data-dependent manner. This flexible scheme empowers the fusion module to concentrate on relevant regions and achieve better-aligned fusion. The noticeable improvements observed on the curved and occluded datasets highlight the significant importance of aligned fusion within our framework.
    
    \subsubsection{Discussion on font properties}

    In this paper, the obtained performance gain can be attributed to the effective utilization of the generated class-aware mask. To comprehensively investigate the relationship between font properties and recognition performance, a series of experiments involving various font types and weights are conducted. By closely examining the fonts present in a subset of test images, we initially select four font types, namely \textit{DIN}, \textit{Times New Roman}, \textit{Arial}, and \textit{Verdana}, for further evaluation. The results presented in Tab.~\ref{tab:font_type} demonstrate that different fonts exert a subtle influence on the final recognition performance. Based on the overall performance across all benchmarks, \textit{Arial} emerges as the preferred font due to its superior performance.

    \begin{table}[t]
        \centering
        \begin{center}
        \caption{Recognition performance with different font types.}
        \label{tab:font_type}
        \scalebox{0.75}{
            \begin{tabular}{c|cccccccccccc|c}
                \toprule
                Font Type & IIIT & SVT  & IC13 & IC15 & SVTP & CUTE & COCO & CTW  & TT   & HOST & WOST & WordArt & Avg.          \\ \midrule
                DIN       & 97.4 & 94.6 & 96.8 & 86.6 & 90.2 & 89.2 & 68.8 & 77.3 & 81.7 & 76.0 & 85.1 & 73.8    & 79.1          \\
                Roman     & 97.3 & 95.2 & 96.8 & 86.8 & 88.4 & 90.3 & 68.8 & 77.4 & 81.8 & 76.7 & 85.5 & 72.9    & \textbf{79.3} \\
                Verdana   & 97.3 & 94.3 & 96.7 & 86.5 & 89.3 & 91.7 & 68.4 & 76.8 & 81.5 & 75.6 & 84.5 & 72.8    & 78.9          \\
                Arial     & 97.3 & 94.6 & 96.7 & 87.0 & 88.8 & 90.6 & 68.7 & 78.4 & 82.1 & 76.2 & 85.1 & 72.9    & \textbf{79.3} \\ \bottomrule
            \end{tabular}
        }
        \end{center}
    \end{table}

    \begin{table}[t]
        \centering
        \begin{center}
        \caption{Recognition performance with different font weights.}
        \label{tab:font_weight}
        \scalebox{0.75}{
            \begin{tabular}{c|cccccccccccc|c}
                \toprule
                Font Weight & IIIT & SVT  & IC13 & IC15 & SVTP & CUTE & COCO & CTW  & TT   & HOST & WOST & WordArt & Avg.          \\ \midrule
                5           & 97.1 & 94.4 & 96.3 & 87.0 & 88.7 & 80.3 & 68.5 & 77.5 & 81.9 & 76.3 & 84.6 & 73.3    & 79.0          \\
                10          & 97.3 & 94.6 & 96.7 & 87.0 & 88.8 & 90.6 & 68.7 & 78.4 & 82.1 & 76.2 & 85.1 & 72.9    & \textbf{79.3} \\
                15          & 97.4 & 94.4 & 97.0 & 87.1 & 89.1 & 89.6 & 68.8 & 78.0 & 81.2 & 75.0 & 85.1 & 74.1    & 79.2          \\
                20          & 97.2 & 94.4 & 96.1 & 86.7 & 89.6 & 90.3 & 68.4 & 77.2 & 81.6 & 77.0 & 84.7 & 72.3    & 79.0          \\ \bottomrule
            \end{tabular}
        }
        \end{center}
    \end{table}
    
    Subsequently, in order to probe the influence of font weight, we examine Tab.~\ref{tab:font_weight}, which clearly illustrates that both extremely thin and excessively thick fonts have a detrimental impact on recognition accuracy. This can be attributed to the challenges posed by thin fonts' fine details, impeding the segmentation network's ability to capture them effectively. Similarly, dense text composed of overly thick fonts may lead to segmentation ambiguity arising from character overlapping. Consequently, a font thickness of 10 is determined to be the optimal choice. Finally, all experiments in this paper are conducted utilizing \textit{Arial} font with a thickness of 10.

    \begin{figure}[t]
        \centering
        \includegraphics[width=0.85\linewidth]{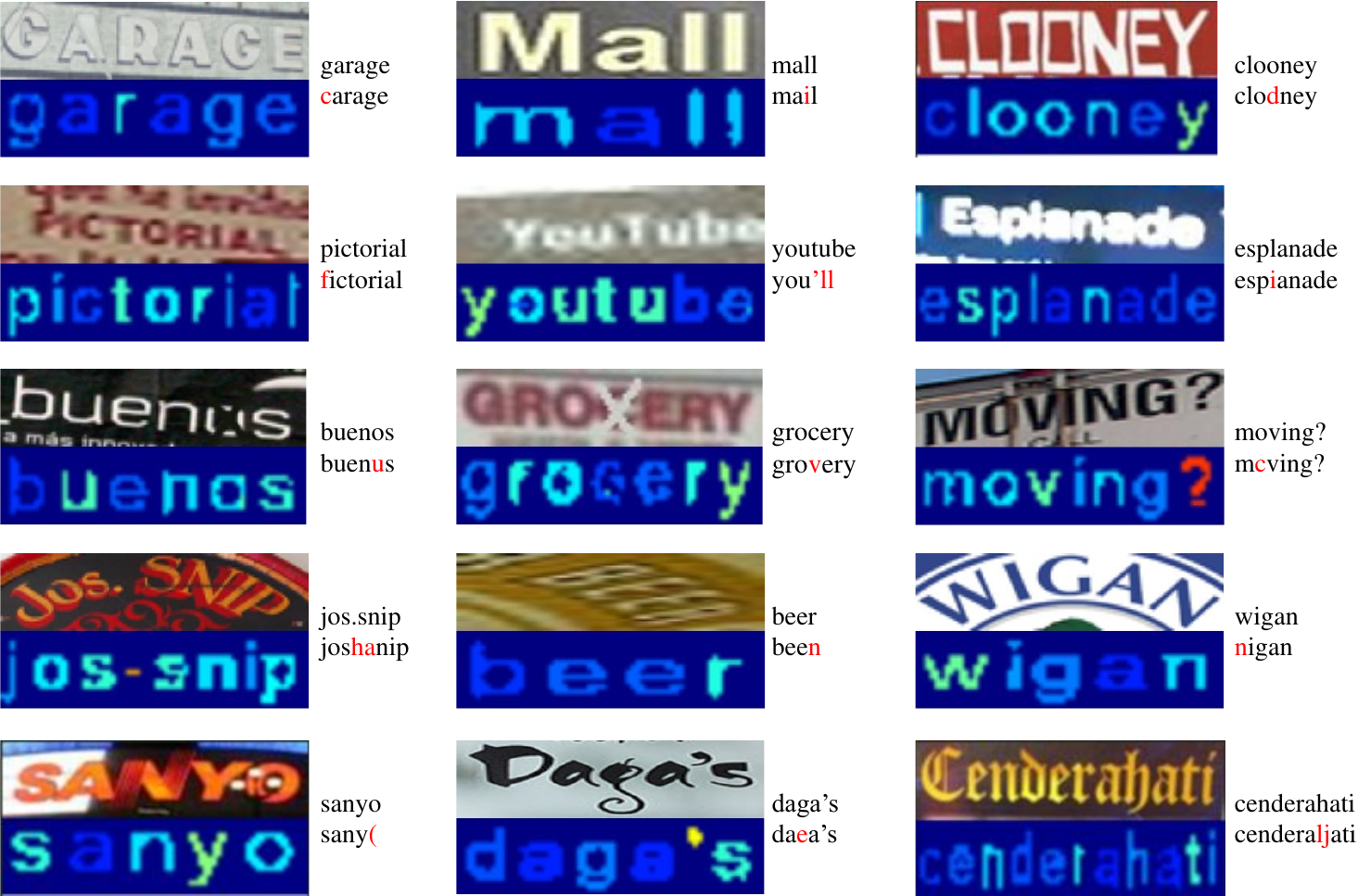}
        \caption{Visualization of the generated canonical class-aware masks in our model without refinement. The two strings near each image represent ground truth and prediction, respectively.}
        \label{fig:vis_binary}
    \end{figure}
    
\subsection{Qualitative Analysis}

    \subsubsection{Complementarity of class-aware masks}

    During our experiments, we initially incorporate the class-aware segmentation branch to implicitly assist text recognition. As observed in Fig.~\ref{fig:vis_binary}, although the final recognition results are incorrect, the class-aware mask can be accurately predicted in certain challenging scenarios, where the mask demonstrates its effectiveness in aligning irregular text to canonical layouts, compensating for incomplete text, and discriminating varied fonts. The generated canonical class-aware masks are generally easier to recognize due to the absence of distortion factors, resulting in more reliable results. However, to avoid cumulative errors, we refrain from directly employing the predicted mask for recognition. Instead, we utilize it to compensate for the original recognition features.
    
    Furthermore, we observe a spatial misalignment between the canonical class-aware mask and the input image. The pixel-wise fusion in this scenario can lead to confusion and adversely impact recognition performance. Therefore, we introduce an aligned feature fusion strategy that improves the integration of the mask and recognition features. It is worth noting that this aligned feature fusion strategy has been overlooked in previous GAN-based or segmentation-based methods, setting our approach apart.
       
    \subsubsection{Character discrimination visualized via t-SNE}

    \begin{figure}[!t]
        \centering
        \includegraphics[width=0.7\linewidth]{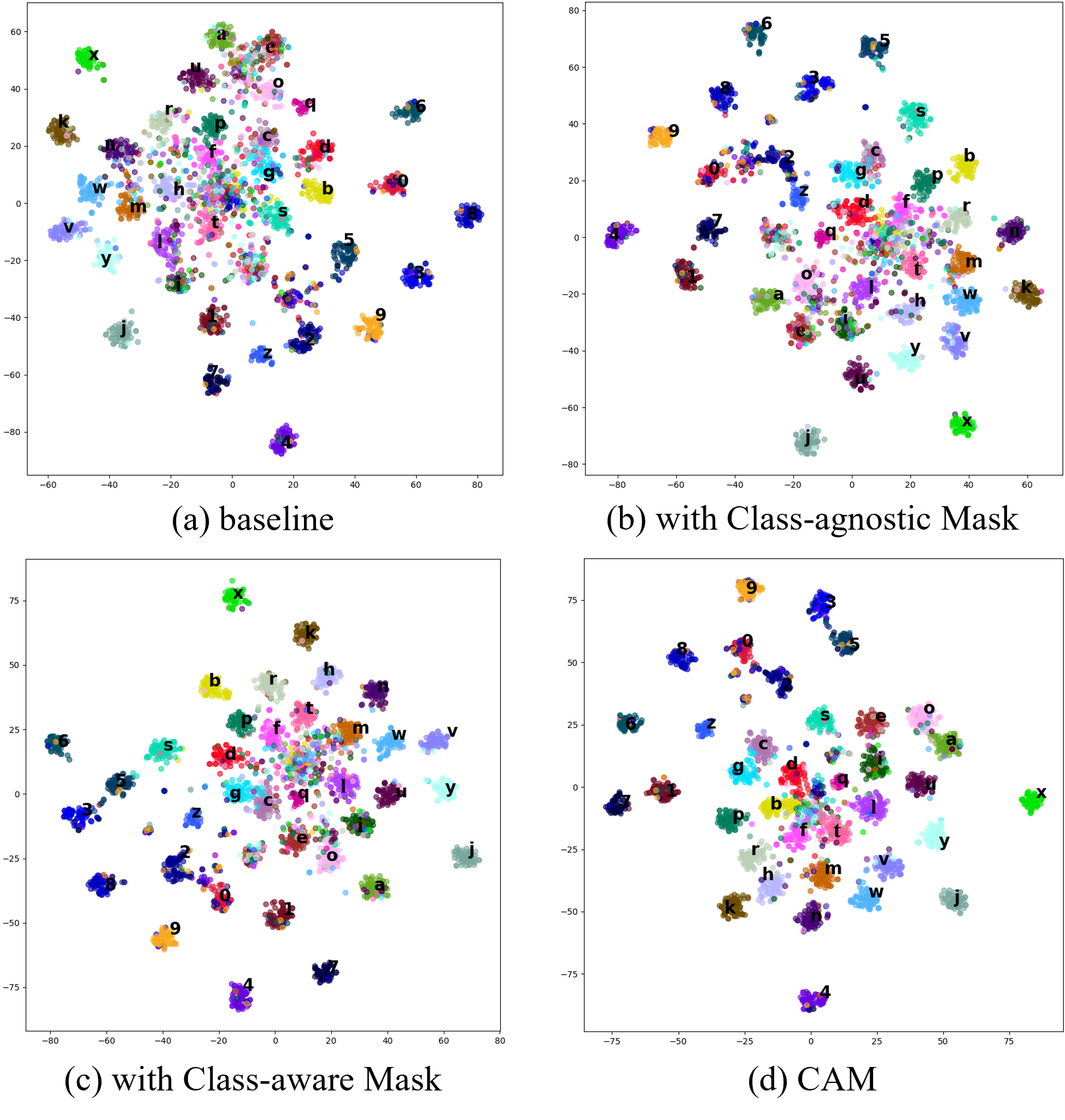}
        \vspace{-4ex}
        \caption{t-SNE visualization of character features, which are extracted from different variants in our study.}
        \vspace{-2ex}
        \label{fig:tsne}
    \end{figure}

    To gain a clearer understanding of the impact of our proposed modules, we conduct further visualization of the character features learned from different model variants. Specifically, we compare the baseline model, the variant with class-agnostic masks, the variant with class-aware masks, and the entire model using t-SNE visualization, as shown in Fig.~\ref{fig:tsne}. The experiments are performed on all test datasets, with 200 images sampled from each dataset. To ensure a legible visualization, the maximum number of sampling points for each character is limited to 100. This analysis provides insights into the distinctive feature representations learned by our model and highlights the effectiveness of our proposed modules.

    The inclusion of our modules results in clearer separation of feature clusters for different characters and reduced scattering of ambiguous cases. This observation highlights the improved robustness and distinguishability of features extracted from various scenarios. Notably, characters with similar shapes, such as `c' and `g' or `G', `2' and `z', `i' and `l', etc., exhibit less entanglement in their feature representations. This indicates that our proposed modules effectively enhance the discrimination of the learned features, enabling better differentiation between characters with similar appearances.
    
    \subsubsection{Visualization of recognition results}

    \begin{figure}
        \centering
        \includegraphics[width=0.8\linewidth]{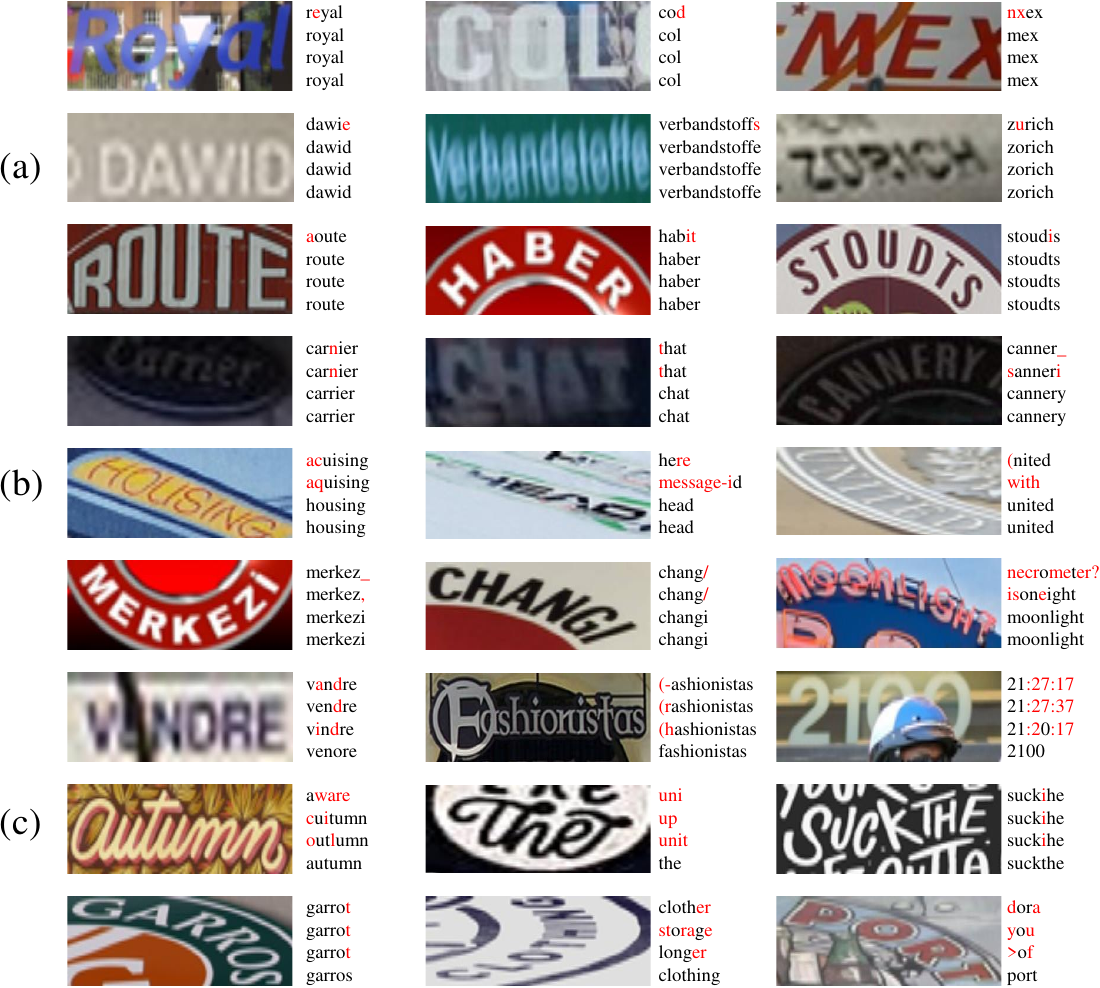}
        \caption{Visualization of text recognition results. The four strings near each image represent the prediction of the baseline model, the variant with class-agnostic masks, the variant with class-aware masks, and our entire model, respectively. (a), (b), (c) are different image groups to demonstrate the effectiveness of each model unit.}
        \label{fig:vis_recognition}
    \end{figure}

    To clearly illustrate the advantages of our proposed modules, we provide qualitative comparisons in Fig.~\ref{fig:vis_recognition}. In group (a), the model variants with mask guidance consistently achieve more robust results in specific scenarios, such as text disturbed by cluttered backgrounds, blur, and curved shapes. The inclusion of the canonical mask into STR helps eliminate style and background noise, as well as handle irregular layouts more effectively.
    
    In group (b), we find that the model benefits from the integration of class-specific information provided by the class-aware mask. This empowers the STR model to tackle challenging situations, including text in low light conditions and heavier perspective distortions and curved shapes. The inclusion of class-specific information allows the model to capture more discriminative representations, resulting in improved text recognition performance.
    
    In group (c), the explicit incorporation of the mask exhibits enhanced robustness in handling various challenging scenarios. Specifically, it effectively addresses word art with larger variances, text instances with severe occlusions, and perspective distortions. The prototype guidance provided by the canonical mask, along with its fusion with the recognition process, offers complementary information that effectively addresses the challenges posed by complex and diverse text instances. 
    
    Finally, the qualitative comparisons underscore the significance of both explicit and implicit utilization of class-aware masks. The incorporation proves effective in handling text instances with complex backgrounds, diverse fonts, flexible arrangements, low image quality, and accidental occlusions.

    \subsubsection{Limitations}

\begin{figure}
        \centering
        \includegraphics[width=0.95\linewidth]{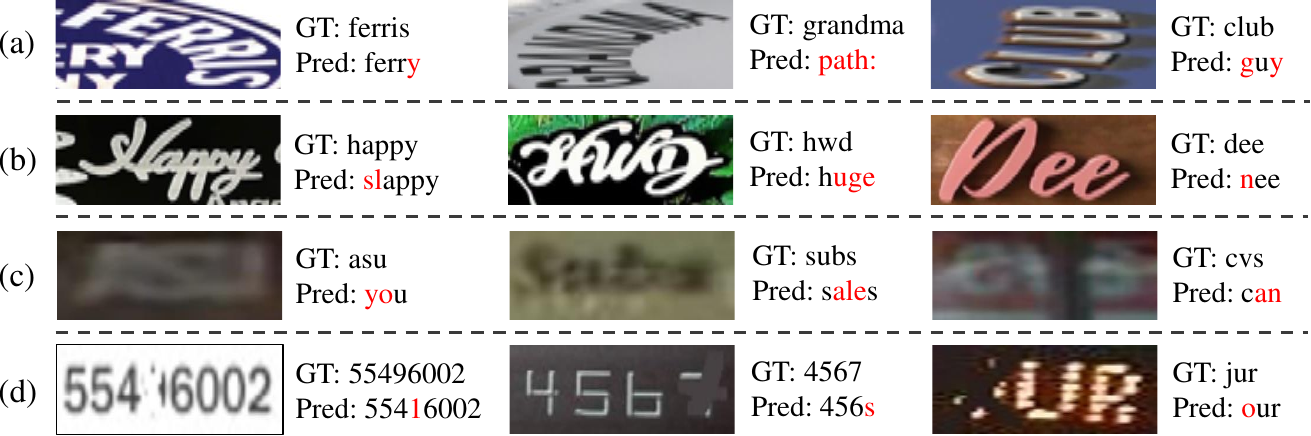}
        \caption{The failure cases of CAM}
        \label{fig:bad_cases}
    \end{figure}

    While CAM can incorporate more discriminative features with the class-aware glyph mask and dynamic fusion, resulting in significant performance improvements, we identify certain limitations in our method. As shown in Fig.~\ref{fig:bad_cases}, CAM struggles with (a) extremely distorted text, (b) text with scare fonts, (c) severely blurred text and (d) heavily occluded text. To address these challenges, we believe that a stronger visual feature representation and the inclusion of language priors would be instrumental.

\section{Conclusion}
    In this paper, we have presented a Class-Aware Mask-guided feature refinement network (CAM) for robust scene text recognition. By leveraging class-aware text masks generated from a standard font, CAM effectively suppresses background and text style noise and thus enhances feature discrimination. Moreover, we have introduced a flexible feature alignment and fusion module to refine the recognition features, further improving text recognition performance by leveraging the complementarity of canonical mask features and recognition features. Extensive evaluations on standard text recognition benchmarks and challenging datasets have validated the effectiveness of our method in aligning irregular text to canonical layouts, compensating for occluded text, and discriminating varied fonts. The entire framework is trained in an end-to-end fashion, achieving a better balance between performance, parameter efficiency, and runtime. We believe the proposed class-aware mask has the potential to inspire peers to explore different representation forms of text labels, while the feature fusion module may offer insights to researchers in the popular multi-modal field. However, CAM still struggles with extremely distorted text, scarce fonts, severely blurred text and heavily occluded text. In future work, we plan to incorporate language priors to address the above challenges and further explore this paradigm for scene text detection and scene text spotting.

\section{Acknowledgments}
This project is supported by the Young Scientists Fund of the National Natural Science Foundation of China (Grant No. 62206103).

\bibliographystyle{elsarticle-num-names} 
\bibliography{references}

\newpage

\end{document}